\documentclass[letterpaper]{article} % DO NOT CHANGE THIS
\usepackage{aaai2026}  % DO NOT CHANGE THIS
\usepackage{times}  % DO NOT CHANGE THIS
\usepackage{helvet}  % DO NOT CHANGE THIS
\usepackage{courier}  % DO NOT CHANGE THIS
\usepackage[hyphens]{url}  % DO NOT CHANGE THIS
\usepackage{graphicx} % DO NOT CHANGE THIS
\usepackage{natbib}  % DO NOT CHANGE THIS AND DO NOT ADD ANY OPTIONS TO IT
\usepackage{caption} % DO NOT CHANGE THIS AND DO NOT ADD ANY OPTIONS TO IT
\usepackage{algorithm}
\usepackage{algorithmic}

\usepackage{multirow}
\usepackage{booktabs}
\usepackage{threeparttable}
\usepackage{siunitx} % 需在导言区加载此宏包

\usepackage{amsmath, amssymb}

\usepackage{newfloat}
\usepackage{listings}
\DeclareCaptionStyle{ruled}{labelfont=normalfont,labelsep=colon,strut=off} % DO NOT CHANGE THIS
\floatstyle{ruled}
\newfloat{listing}{tb}{lst}{}
\floatname{listing}{Listing}
\title{\textit{\underline{R}}ecall-\textit{\underline{E}}xtend \textit{\underline{D}}ynamics: Enhancing Small Language Models through \\ Controlled Exploration and Refined Offline Integration}
\author{
    Zhong Guan\textsuperscript{\rm 1},
    Likang Wu\textsuperscript{\rm 1},
    Hongke Zhao\textsuperscript{\rm 1},
    Jiahui Wang\textsuperscript{\rm 1},
    Le Wu\textsuperscript{\rm 2}
}
\affiliations{
    \textsuperscript{\rm 1}Tianjin University\\
    \textsuperscript{\rm 2}Hefei University of Technology \\
}

\usepackage{bibentry}
\begin{document}

\maketitle

\begin{abstract}
Many existing studies have achieved significant improvements in the reasoning capabilities of large language models (LLMs) through reinforcement learning with verifiable rewards (RLVR), while the enhancement of reasoning abilities in small language models (SLMs) has not yet been sufficiently explored. Combining distilled data from larger models with RLVR on small models themselves is a natural approach, but it still faces various challenges and issues. Therefore, we propose \textit{\underline{R}}ecall-\textit{\underline{E}}xtend \textit{\underline{D}}ynamics(RED): Enhancing Small Language Models through Controlled Exploration and Refined Offline Integration. In this paper, we explore the perspective of varying exploration spaces, balancing offline distillation with online reinforcement learning. Simultaneously, we specifically design and optimize for the insertion problem within offline data. By monitoring the ratio of entropy changes in the model concerning offline and online data, we regulate the weight of offline-SFT, thereby addressing the issues of insufficient exploration space in small models and the redundancy and complexity during the distillation process. Furthermore, to tackle the distribution discrepancies between offline data and the current policy, we design a sample-accuracy-based policy shift mechanism that dynamically chooses between imitating offline distilled data and learning from its own policy.
\end{abstract}

\begin{links}
    \link{Code}{https://github.com/millioniron/OpenRLHF-Millioniron-/tree/master}
\end{links}

\section{Introduction}

Recent studies~\cite{jaech2024openai,guo2025deepseek,team2025kimi} indicate that  LLMs can achieve significant performance improvements in various tasks through enhanced reasoning capabilities. A key technology driving this progress is  RLVR~\cite{lambert2024tulu,shao2024deepseekmath}, which has garnered widespread attention in the research community and led to the development of various policy optimization algorithms such as GRPO~\cite{guo2025deepseek}, DAPO~\cite{yu2025dapo}, GPG~\cite{chu2025gpg}, and DR.GRPO~\cite{liu2025understanding}, and others~\cite{chen2025minimax,zhang2025gvpo}. These methods are characterized by their simplicity, directness, and result-oriented approach, distinguishing them from traditional methods based on process reward modeling~\cite{liprocess,setlur2024rewarding}.

Despite the promising advancements in the aforementioned reasoning-related research, most studies have focused on models ranging from 7B to 32B in parameters. For smaller models (e.g., 1.5B), research has primarily  adopted direct or multi-stage distillation methods~\cite{yang2025qwen3,guo2025deepseek,abdin2025phi}, and their integration with RLVR remains largely unexplored.

\begin{figure}
    \centering
    \includegraphics[width=1\linewidth]{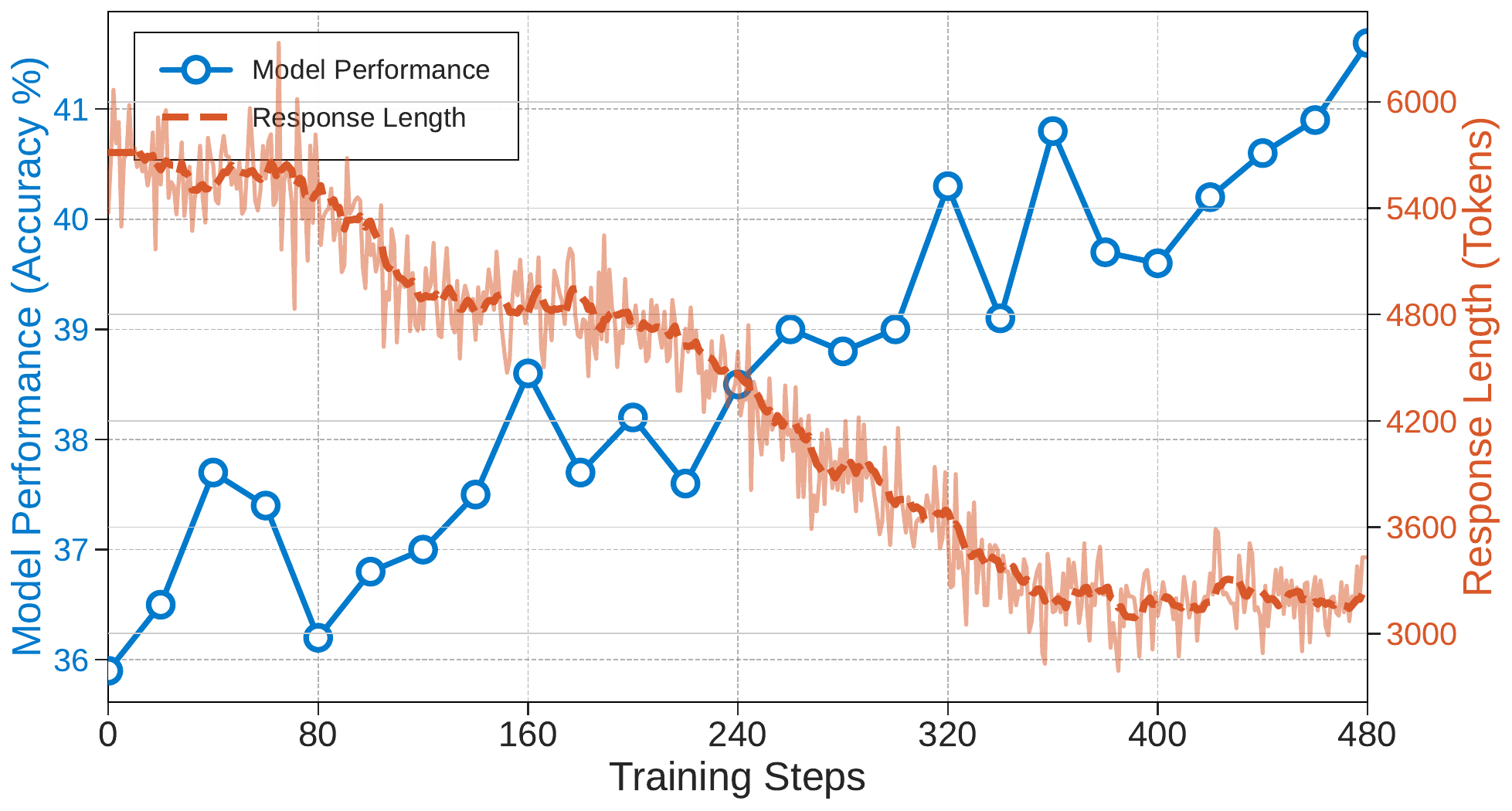}
    \caption{RL training on an SFT-distilled model. The pre-SFT model overthinks significantly, generating a lot of redundant content that leads to inefficient training. There's still room to improve the model's performance.}
    \label{fig:figure0}
\end{figure}

However, small models trained solely through Supervised Fine-Tuning (SFT) distillation still exhibit noticeable issues such as overthinking and redundant generation~\cite{zhang2025making,li2025small}, indicating room for overall performance improvement. As illustrated  in the Figure~\ref{fig:figure0}, pre-SFT models optimized with post-RL show improved reasoning performance and can, to some extent, reduce the probability of generating overthinking words, thereby decreasing redundant output.

% \begin{figure}
%     \centering
%     \includegraphics[width=1\linewidth]{figure2.png}
%     \caption{figure2}
%     \label{fig:figure2}
% \end{figure}

Nevertheless, the ``pre-SFT + post-RL" training pipeline still faces efficiency challenges. As depicted in the Figure~\ref{fig:figure0}, the model's generation length during RL training gradually decreases from an initial over 6,000 tokens to approximately 3,000, while the rollout time, influenced by generation length, significantly increases, further prolonging the training cycle.

Therefore, this paper proposes a more efficient strategy combining offline-SFT with online-RL, aiming to enhance the reasoning capabilities of small-scale models. Compared to existing SFT + RL methods~\cite{yan2025learning,fu2025srft,zhang2025bread,huang2025blending}, our innovation lies in exploring changes in the exploration space to balance offline distillation data with online reinforcement learning, alongside  specifically designing and optimizing for the integration problem in offline data.

(I) Controlled  Exploration: Balancing Recall and Extend  via \textbf{Dynamic Entropy Regulation}

Existing research has confirmed that RLVR, in most cases, does not endow models with new capabilities but rather acts as an ``amplifier", uncovering and activating existing knowledge from the model's pre-training phase~\cite{yue2025does,shao2024deepseekmath}. Specifically, RLVR demonstrates performance improvement for smaller $k$ in pass@$k$, but its effect declines for larger $k$. We define this process as a \textbf{Recall} phase, where the core objective is to optimize the reasoning path within existing capabilities while contracting the exploration space. SFT, conversely, can expand the reasoning boundary, introducing new reasoning patterns learned from stronger teacher models and increasing the model's explorable space, thus seen as an \textbf{Extend} of the model's capabilities~\cite{guo2025deepseek,kim2025reinforcement}.

These two approaches can complement each other to some extent. We propose using the entropy variation ratio, an intuitive metric for explorable space, to balance the Recall and Extend phases, addressing the issues of insufficient RL exploration space in small models and the redundancy and complexity problems inherent in the distillation process, thereby achieving effective complementarity between the two.

(II) Adaptive Integration of Offline Data with Accuracy-aware Policy Shifts

Integrating distilled offline data into the policy optimization function can achieve a unified training paradigm. However, using a fixed clip value makes it difficult to handle distillation data that often deviates significantly from the current policy. Furthermore, if distillation samples are treated as a deterministic policy (i.e., setting $\pi_{\text{offline}}=1$), it can rapidly exacerbate entropy collapse; conversely, directly integrating distillation samples via SFT loss or an on-policy form may lead to model performance collapse. 

To address these issues, we propose a method that dynamically adjusts the policy offset based on the answer's correctness rate. For samples with high correctness rates, we prefer setting $\pi_{\text{offline}}=1$, allowing the model to learn from its own policy; for samples with low correctness rates, the policy offset is set closer to $\pi$, encouraging  the model more inclined to imitate the distillation samples. This method not only improves the model's adaptability and robustness to data of varying quality but also optimizes overall training efficiency. In brief, our contributions can be summarized as follows:

\begin{itemize}
\item We frame the integration of RL and SFT as a synergy process of Recall and Extend. By dynamically balancing  these two stages, we enable Extend phase to satisfy the exploration space required by Recall phase, while Recall refines and simplifies the complex information from Extend. This achieves steady improvement in smaller models using offline distilled data.
\item Our approach significantly improves integration of offline distilled data into the RL process. We introduced a policy offset term based on Accuracy-aware to estimate $\pi_{\text{offline}}$. This mechanism allows the model to adjust autonomously when processing samples with high accuracy, while leaning more towards imitating the large model's strategy for samples with high error rates. This effectively avoids entropy collapse and performance degradation.
\end{itemize}

\section{Related Work}

\subsection{Reasoning with Integrated SFT and RL}

UFT~\cite{wang2024uft} proposes a new method that combines the training paradigms of DPO~\cite{rafailov2023direct} and SFT, directly aligning the implicit rewards under the DPO's Sigmoid functio with true labels. NFT~\cite{chen2025bridging} demonstrates  that by additionally utilizing negative samples, the performance gap between SFT and mainstream RL algorithms can be significantly narrowed. In ReLIFT~\cite{ma2025learning}, researchers collected difficult problems during the RL phase and conducted targeted fine-tuning during the SFT phase, effectively combining RL and SFT to enhance LLM's reasoning and out-of-distribution generalization capabilities, achieving performance beyond the model's inherent cognitive constraints. SASR~\cite{chen2025step} proposes an adaptive hybrid training framework that dynamically selects SFT or RL training strategies based on the gradients of SFT and GRPO tasks, achieving a dynamic balance between the two throughout the optimization process. BREAD~\cite{zhang2025bread} points out that the inherent performance of small models affects the synergistic effect of SFT and RL. This method introduces branch rollout to coordinate the two-stage training; for completely failed problems, concise expert prefixes or prompts will be adaptively inserted in the next stage to guide the small model to complete the remaining reasoning path. At a more fundamental level, LUFFY~\cite{yan2025learning} introduces distillation data from strong reasoning models, modeling it as off-policy guidance, unified within a zero-RL training paradigm, and integrating the advantages of SFT and RL in a mix-policy form. Furthermore, SRFT~\cite{fu2025srft}, to more effectively utilize SFT for example learning and RL for policy exploration, proposes an entropy-aware weighting mechanism, organically combining the two fine-tuning paradigms to construct a unified training framework.

\subsection{Small Model Reasoning}

\begin{figure*}[t]
    \centering
    \includegraphics[width=1\linewidth]{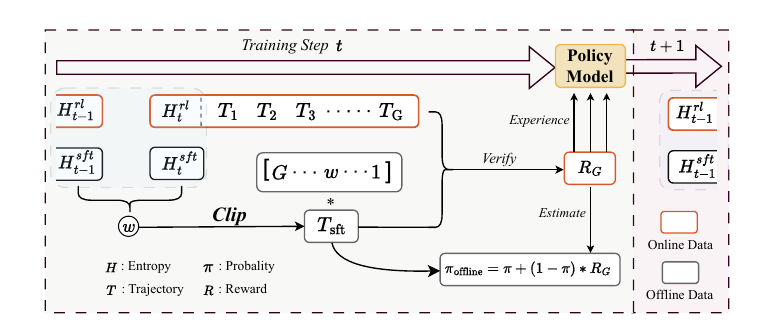}
    \caption{The training pipeline of RED. The training pipeline of RED. RED computes the weight $w$ for SFT based on the entropy dynamics of RL and SFT, thereby balancing the contributions of SFT and RL. Additionally, the off-policy probability $\pi_{\text{offline}}$ is determined according to the sample accuracy — the lower the accuracy, the higher the uncertainty, and vice versa.}
    \label{fig:figure2}
\end{figure*}

Early community efforts primarily adopted direct distillation to endow small models with reasoning capabilities~\cite{yang2025qwen3,guo2025deepseek}. Furthermore, \cite{xu2025phi}proposed a systematic training scheme for SLMs, improving small model performance through a hybrid of SFT and reinforcement learning (RL) stages. ADPA\cite{gaoadvantage} optimizes the student model's policy by estimating the advantage function between a trained preference-aligned teacher model and a reference model, providing distribution-level reward signals. REDI~\cite{xu2025harnessing} introduces negative example traces to enhance reasoning distillation, proposing an asymmetrically weighted reference-free objective, which brings performance gains without expensive online reinforcement learning interactions. \cite{chen2025towards} further proposed a re-distillation strategy, generating new SFT training data by sampling from a converged policy to improve model capability. \cite{kim2025reinforcement}'s research found that while distillation can improve accuracy, its true potential for raising the model's upper capacity limit depends on introducing new knowledge. If only existing reasoning patterns are distilled, similar to RLVR, it may instead reduce accuracy for some problems. \cite{zhang2025making} found in extensive experiments that SFT-based distillation methods lead to severe inefficiency in small models. This inefficiency primarily stems from redundant generation and repetitive content, especially under increased computational budgets during testing, resulting in non-monotonic improvements in accuracy.

\section{Preliminary}

In standard GRPO, for each query $q \sim \mathcal{D}$, the model samples a set of responses $O = \{ o_1, o_2, \cdot \cdot \cdot, o_G \}$ from the old policy $\pi_{\theta_{\text{old}}}$, and optimizes the policy model $\pi_{\theta}$ by maximizing the following objective:
\begin{equation}
% \small
\begin{aligned}
\mathcal{J}(\theta) =\;&
\mathbb{E}_{(q,a) \sim \mathcal{D},\, \{o_i\}_{i=1}^{G} \sim \pi_{\theta}(\cdot|q)} \left[
\frac{1}{G} \sum_{i=1}^{G} \frac{1}{|o_i|} \sum_{t=1}^{|o_i|}
\right. \\
& \left.
\left\{
\min\left[ r_{i,t}(\theta)\hat{A}_{i,t},\,
\text{clip}\left(r_{i,t}(\theta), 1\!-\!\epsilon, 1\!+\!\epsilon \right)\hat{A}_{i,t} \right]
\right. \right. \\ % Break here
& \quad \left. \left.
- \beta\, \mathbb{D}_{\text{KL}}\left[\pi_\theta \,\middle\|\, \pi_{\text{ref}} \right]
\right\}
\right].
\end{aligned}
\end{equation}

Here, $r_{i,t}(\theta)$ is the policy ratio at token $t$ of response $o_i$, $\hat{A}_{i,t}$ is the advantage estimate, and $\epsilon$ controls the clipping range.

Let us further simplify the expression by removing the KL divergence term and assuming that $r_{i,t}(\theta)$ is always within the clipping range.
\begin{equation}
\mathcal{J}(\theta) = \mathbb{E}_{(q,a) \sim \mathcal{D},\, \{o_i\}_{i=1}^{G} \sim \pi_{\theta}(\,\cdot\,|q)} \left[ \frac{1}{G} \sum_{i=1}^{G} \frac{1}{|o_i|} \sum_{t=1}^{|o_i|} r_{i,t}(\theta)\hat{A}_{i,t} \right].
\end{equation}

Taking the gradient of the objective function $\mathcal{J}(\theta)$ yields:
\begin{equation}
\small
\begin{aligned}
\nabla \mathcal{J(\theta)} &= \mathbb{E}_{{(q,a) \sim \mathcal{D}},\{o_i\}_{i=1}^{G} \sim \pi_{\theta}(.|q) } \left\{ \frac{1}{G}\sum_{i=1}^{G} \frac{1}{|o_i|} \sum_{t=1}^{|o_i|} \nabla r_{i,t}(\theta)\hat{A}_{i,t}\right\}\\
&= \left\{ \frac{1}{G}\sum_{i=1}^{G} \frac{1}{|o_i|} \sum_{t=1}^{|o_i|} r_{i,t} (\theta) \hat{A}_{i,t} \nabla \log \pi_{\theta}(o_{i,t}|q,o_{i,<t} ) \right\}.
\end{aligned}
\end{equation}

When performing SFT on a small model using distilled reasoning data $(q, o_{\text{sft}}) \sim \mathcal{D}_{\text{sft}}$, the loss function takes the form of a maximum likelihood objective. Supervised learning fundamentally aims to train a model $\pi_\theta(o\mid q)$ to approximate the underlying data distribution $\pi_{sft}(o^{sft}\mid q)$. This can be achieved by optimizing the following maximum likelihood objective:
\begin{equation}
\small
\nabla \mathcal{L}_{\text{sft}}(\theta) = \mathbb{E}_{(q, o^{sft}) \sim \mathcal{D}_{\text{sft}}} \Bigg\{ 
 \frac{1}{|o^{sft}|}\sum_{t=1}^{|o|} \nabla\log \pi_{\theta}(o_t \mid q, o_{< t}) \Bigg\}.
\end{equation}

In the post-training stage, a simple loss function form of the training framework that unifies RL and SFT can be written as:
% \begin{equation}
% \label{eq:eq5}
% \nabla\mathcal{J_{{\text{mix}}} (\theta)} = \mathbb{E}_{(q, o_{sft}) \sim \mathcal{D}_{\text{sft}}} \Bigg\{ 
% \frac{1}{G}\sum_{i=1}^{G} \frac{1}{|o_i|} \sum_{t=1}^{|o_i|} r_{i,t}(\theta)\hat{A}_{i,t} \nabla\log \pi_{\theta}(o_t \mid q, o_{< t})+\frac{1}{|o|}\sum_{t=1}^{|o|} \nabla\log \pi_{\theta}(o_t \mid q, o_{< t}) \Bigg\}
% \end{equation}

% \begin{equation}
% \label{eq:eq5}
% \begin{aligned}
% \nabla \mathcal{J_{\text{mix}}(\theta)} &= \mathbb{E}_{{(q,a) \sim \mathcal{D}},\{o_i\}_{i=1}^{G} \sim \pi_{\theta}(.|q)}\Bigg\{ & \frac{1}{G}\sum_{i=1}^{G} \frac{1}{|o_i|} \sum_{t=1}^{|o_i|} r_{i,t}(\theta)\hat{A}_{i,t} \nabla\log \pi_{\theta}(o_t \mid q, o_{< t}) \\
% & + \frac{1}{|o|}\sum_{t=1}^{|o|}\nabla\log \pi_{\theta}(o_t \mid q, o_{< t}) \Bigg\}
% \end{aligned}
% \end{equation}
\begin{equation}
% \small
\label{eq:eq5}
\begin{aligned}
&\nabla \mathcal{J}_{\text{mix}}(\theta) 
= \mathbb{E}_{(q,o^{sft}) \sim \mathcal{D},\{o_i\}_{i=1}^{G} \sim \pi_{\theta}(.|q)} \\
& \Bigg\{ \frac{1}{G}\sum_{i=1}^{G} \frac{1}{|o_i|} \sum_{t=1}^{|o_i|} r_{i,t}(\theta)\hat{A}_{i,t} \nabla\log \pi_{\theta}(o_t \mid q, o_{< t}) \\
&\quad + \frac{1}{|o^{sft}|}\sum_{t=1}^{|o^{sft}|}\nabla\log \pi_{\theta}(o^{sft}_t \mid q, o_{< t}) \Bigg\}.
\end{aligned}
\end{equation}

\section{Method}

In this section, we provide a detailed description of our approach, which is divided into two main parts: (I)Balancing Recall and Extend  via Dynamic Entropy Regulation, and (II)Adaptive Integration of Offline Data with Accuracy-aware Policy Shifts.

\subsection{Adaptive Integration of Distilled Data}

\begin{figure}
    \centering
    \includegraphics[width=1\linewidth]{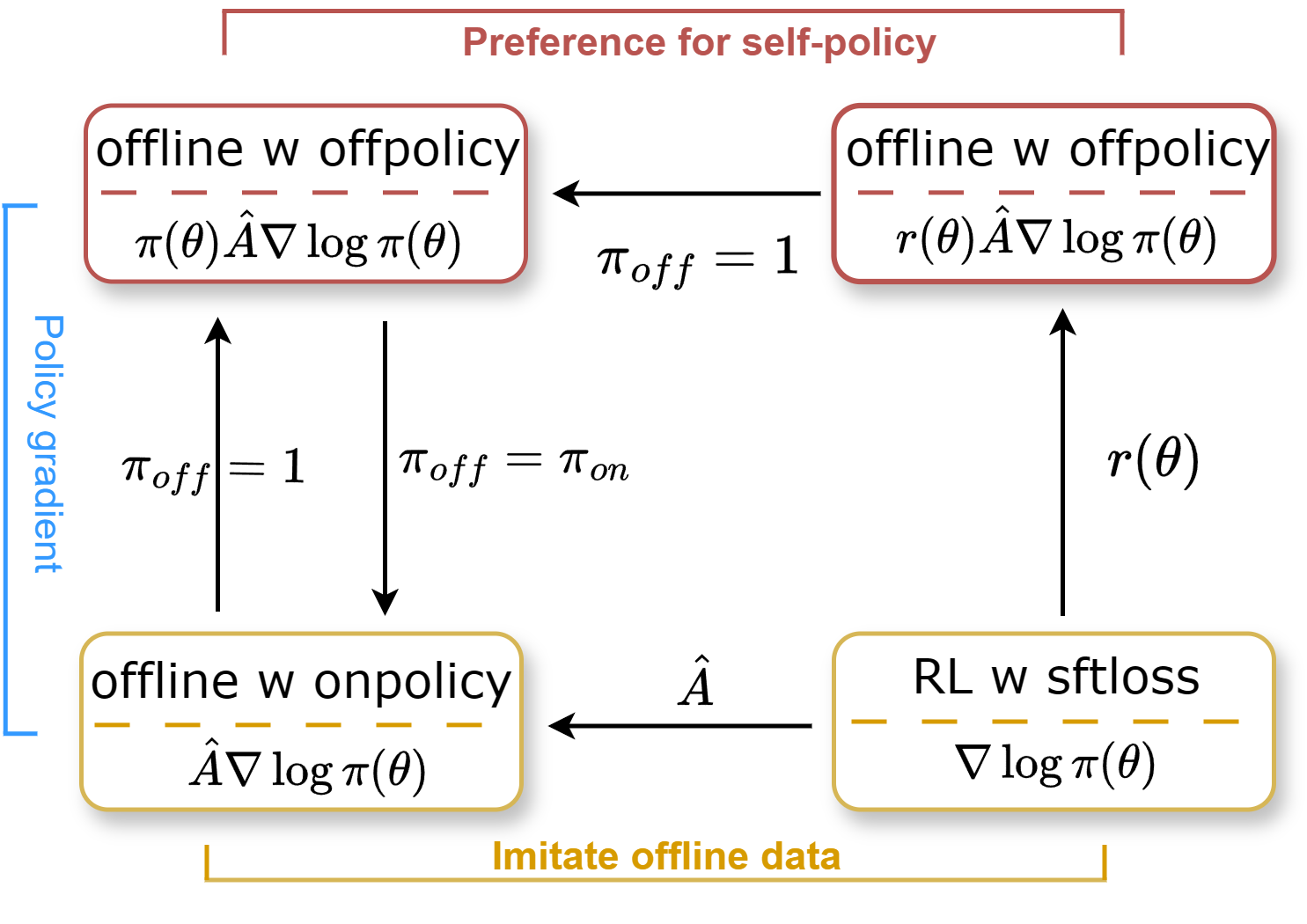}
    \caption{ Transformations Between Different Forms of Offline-SFT Integrated into the RL Policy Function. This figure illustrates the  conditions for converting between various forms when integrating offline-SFT into the RL policy function.}
    \label{fig:figure3}
\end{figure}

To effectively leverage the advantages of offline-SFT and online-RL, it is essential to integrate their respective objective (loss) functions. In the Preliminary, Eq(\ref{eq:eq5}) presents a simple fusion of RL and SFT. Building on this, an intuitive solution is to incorporate the generated trajectory of the distilled model into the policy optimization function. Furthermore, we illustrate the relationship between these strategies in  Figure~\ref{fig:figure3}. By adding an  advantage term, RL with SFT loss can be easily transformed into on-policy format, 
\begin{equation}
% \small
\label{eq:eq6}
\begin{aligned}
&\nabla \mathcal{J}_{\text{mix}}(\theta) 
= \mathbb{E}_{(q,o^{sft}) \sim \mathcal{D},\{o_i\}_{i=1}^{G} \sim \pi_{\theta}(.|q)} \\
& \Bigg\{ \frac{1}{G+1}\sum_{i=1}^{G} \frac{1}{|o_i|} \sum_{t=1}^{|o_i|} r_{i,t}(\theta)\hat{A}_{i,t} \nabla\log \pi_{\theta}(o_t \mid q, o_{< t}) \\
&\quad + \frac{1}{G+1}\frac{1}{|o^{sft}|}\sum_{t=1}^{|o^{sft}|}\hat{A}_{i,t}\nabla\log \pi_{\theta}(o^{sft}_t \mid q, o_{< t}) \Bigg\}.
\end{aligned}
\end{equation}

And further incorporating importance sampling results in an off-policy formulation.
\begin{equation}
% \small
\label{eq:eq7}
\begin{aligned}
&\nabla \mathcal{J}_{\text{mix}}(\theta) 
= \mathbb{E}_{(q,o^{sft}) \sim \mathcal{D},\{o_i\}_{i=1}^{G} \sim \pi_{\theta}(.|q)} \\
& \Bigg\{ \frac{1}{G+1}\sum_{i=1}^{G} \frac{1}{|o_i|} \sum_{t=1}^{|o_i|} r_{i,t}(\theta)\hat{A}_{i,t} \nabla\log \pi_{\theta}(o_t \mid q, o_{< t}) \\
&\quad + \frac{1}{G+1}\frac{1}{|o^{sft}|}\sum_{t=1}^{|o^{sft}|}r^{\text{offline}}_{i,t}\hat{A}_{i,t}\nabla\log \pi_{\theta}(o^{sft}_t \mid q, o_{< t}) \Bigg\},
\end{aligned}
\end{equation}

where \[r^{\text{offline}}_{i,t}=\frac{\pi}{\pi^{\text{offline}}}.\]

Here, $\pi^{\text{offline}}$ denotes the probability assigned by the larger model's probability for the current trajectory. However, obtaining $\pi^{\text{offline}}$ from a large model is costly, especially with large-scale distillation data. Furthermore, due to inconsistencies in model architectures and differences in vocabularies between large and small models, acquiring an accurate $\pi^{\text{offline}}$ is challenging.

\begin{figure}
    \centering
    \includegraphics[width=1\linewidth]{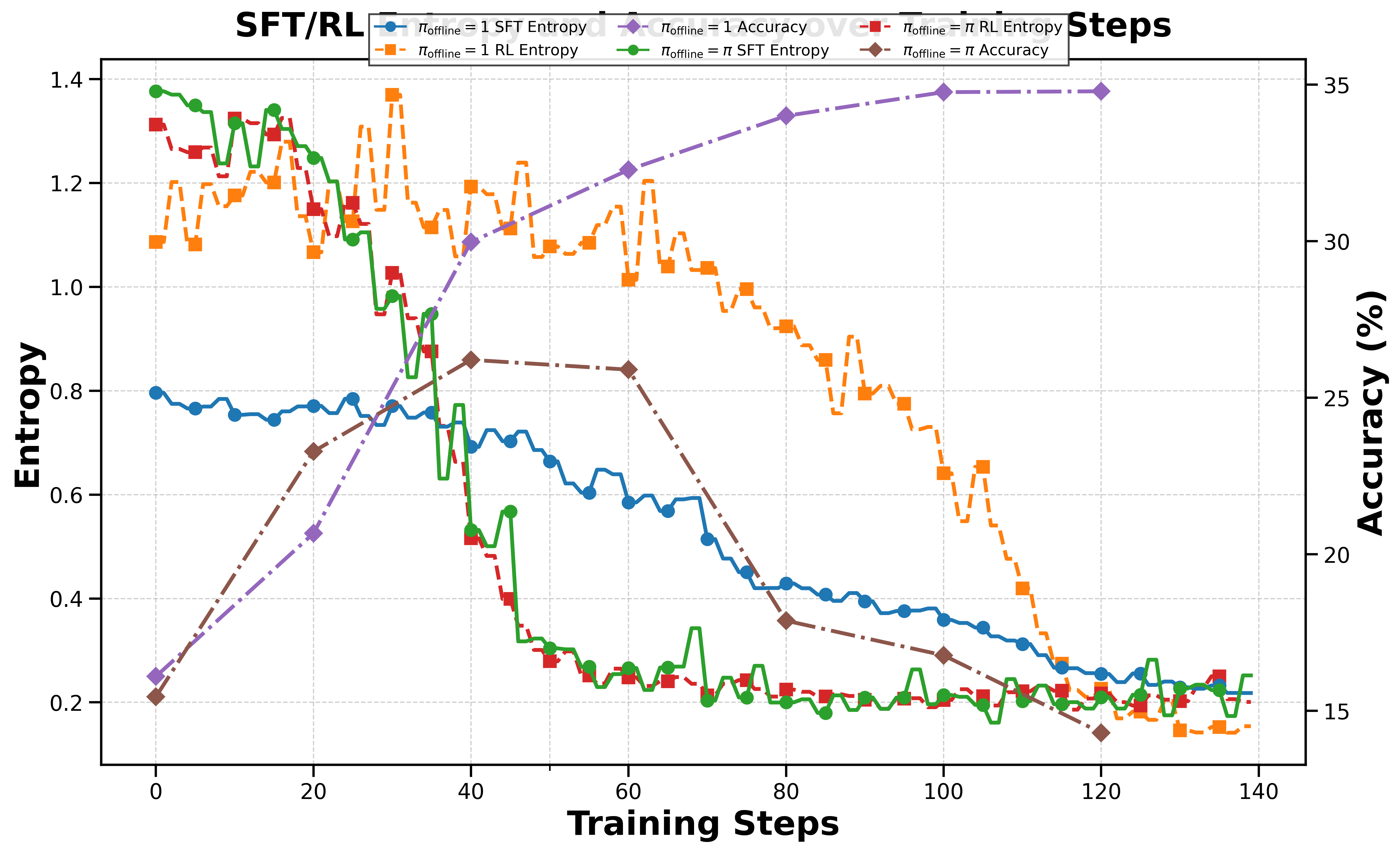}
    \caption{SFT/RL Entropy and Accuracy over Training Steps. This figure illustrates the model's behavior during training when $\pi^{\text{offline}}=1$ and when $\pi^{\text{offline}}=\pi$}
    \label{fig:figure4}
\end{figure}

Conversely, using a naive assignment for $\pi^{\text{offline}}$ leads to various issues. When $\pi^{\text{offline}}=1$, the model's learning from offline data effectively defaults to relying excessively on its own policy($r^{\text{offline}}_{i,t}=\pi$), This causes both SFT entropy and RL entropy to collapse rapidly, as shown in Figure~\ref{fig:figure4}. When $\pi^{\text{offline}}$ is switched to $\pi$, the model treats offline trajectories as on-policy, this offline-onpolicy approach causes model performance to collapse in the mid-training phase. This demonstrates that the value of $\pi^{\text{offline}}$ is crucial for the integration  of offline-SFT.

To address this issue, we introduce a policy deviation term based on problem difficulty to estimate $\pi^{\text{offline}}$. Intuitively, the higher the problem's difficulty, the greater its uncertainty, and thus the smaller the value of $\pi$. Therefore, we added the following deviation term:
\begin{equation}
\label{eq:eq8}
\pi^{\text{offline}}=\pi+(1-\pi)*r_{\text{mean-group}}.
\end{equation}

\begin{table}[h]
\centering
\begin{tabular}{@{}llSSSS@{}}
\toprule
\textbf{Training Phase} & \textbf{Initial Model} & 
\multicolumn{2}{c}{\textbf{Step 0}} & 
\multicolumn{2}{c}{\textbf{Step 300}} \\
\cmidrule(lr){3-4} \cmidrule(l){5-6}
& & {RL} & {SFT} & {RL} & {SFT} \\
\midrule

\multirow{4}{*}{\textbf{RL Phase}} 
& Base & 1.20 & 1.05 & 0.22 & 0.99 \\
& SFT-50 & 0.88 & 0.67 & 0.35 & 0.65 \\
& SFT-100 & 0.56 & 0.44 & 0.30 & 0.43 \\
& SFT-300 & 0.47 & 0.34 & 0.32 & 0.34 \\

\midrule

\multirow{4}{*}{\textbf{SFT Phase}} 
& Base & 1.20 & 1.05 & 0.47 & 0.34 \\
& RL-50 & 0.44 & 0.90 & 0.44 & 0.33 \\
& RL-100 & 0.36 & 0.89 & 0.45 & 0.30 \\
& RL-300 & 0.30 & 0.88 & 0.44 & 0.30 \\
\bottomrule
\end{tabular}

\caption{Entropy evolution across training phases and model checkpoints}
\label{tab:entropy_models_steps}

\end{table}

For samples with high accuracy, we tend to set $\pi^{\text{offline}}=1$, allowing the model to adapt based on its own circumstances. Conversely, for samples with high error rates, $\pi^{\text{offline}}$ tends toward $\pi$, making the model more inclined to imitate the large model's policy.

\subsection{Controlled  Exploration: Balancing Recall and Extend via Dynamic Entropy Regulation}

Our direct motivation stems from the observation that when the exploration space of RLVR is insufficient or incapable of addressing current problems, SFT should be introduced to expand the effective exploration space—and this expansion should occur gradually.

During SFT+RL training, RLVR serves as a Recall mechanism, refining reasoning paths within the model's current capabilities while reducing the exploration space. In contrast, SFT acts as an Extend mechanism, enabling the model to acquire out-of-domain knowledge and increase the range of exploration.

By monitoring the changes in the exploration space( Entropy :$H= -\sum_{i=1}^{|V|} p_i \log p_i$), 
we dynamically adjust the contributions of offline-SFT and RLVR, aiming to balance the reduction in exploration from Recall and the expansion from Extend. Unlike prior work that directly uses entropy values as weighting signals, we instead use the change in entropy from RL stages as an indicator of exploration dynamics($\Delta H^{rl}=|1-\frac{H^{rl}_{t}}{H^{rl}_{t-1}}|,\Delta H^{sft}=|1-\frac{H^{sft}_{t}}{H^{sft}_{t-1}}|$).

When the change in RL entropy is small, we increase the contribution of offline-SFT to boost the exploration space and improve the model’s performance ceiling. Conversely, when RL entropy exhibits sufficient change (indicating active exploration), the influence of offline-SFT is reduced. We further incorporate changes in SFT entropy to balance the emphasis between RL and SFT. Specifically, 
\begin{equation}
\label{eq:eq9}
w=\frac{\Delta H^{sft}}{\Delta H^{rl}},
\end{equation}

It is noted that, as shown in the Table~\ref{tab:entropy_models_steps}, as training progresses, RL training leads to a decrease in RL entropy but does not cause a change in SFT entropy. In contrast, normal SFT causes changes in both SFT entropy and RL entropy. Therefore, by solely adjusting the weight for offline-SFT, we can achieve fine-grained control over the exploration space.

\begin{equation}
% \small
\label{eq:eq10}
\begin{aligned}
&\nabla \mathcal{J}_{\text{mix}}(\theta) 
= \mathbb{E}_{(q,o^{sft}) \sim \mathcal{D},\{o_i\}_{i=1}^{G} \sim \pi_{\theta}(.|q)} \\
& \Bigg\{ \frac{1}{G+1}\sum_{i=1}^{G} \frac{1}{|o_i|} \sum_{t=1}^{|o_i|} r_{i,t}(\theta)\hat{A}_{i,t} \nabla\log \pi_{\theta}(o_t \mid q, o_{< t}) \\
&\quad + \frac{w}{G+1}\frac{1}{|o^{sft}|}\sum_{t=1}^{|o^{sft}|}r^{\text{offline}}_{i,t}\hat{A}_{i,t}\nabla\log \pi_{\theta}(o^{sft}_t \mid q, o_{< t}) \Bigg\}.
\end{aligned}
\end{equation}

To ensure numerical stability and interpretability, we apply an upper and lower clip to the computed weight within the range $[1, G]$, where $G$ is the number of samples in a group. When the weight is 1, it implies that the offline-SFT update acts on a single sample within the group; when it equals $G$, it indicates a full parallel fusion of SFT and RL.

\begin{equation}
\label{eq:eq11}
w=\text{clip}(\frac{\Delta H^{sft}}{\Delta H^{rl}},1,G).
\end{equation}

By employing dynamic entropy regulation, we effectively combine RL to refine reasoning with SFT to broaden the exploration space.

\begin{table*}[tbp]
    \centering

    \begin{tabular}{lcccccccccccccc}
        \toprule
        \multirow{2}{*}{\textbf{Model}} &
        \multicolumn{2}{c}{\textbf{AIME24}} &
        \multicolumn{2}{c}{\textbf{AIME25}} &
        \multicolumn{2}{c}{\textbf{AMC}} &
        \multicolumn{2}{c}{\textbf{MATH500}} &
        \multicolumn{2}{c}{\textbf{Minerva}} &
        \multicolumn{2}{c}{\textbf{Olympiad}} &
        \multicolumn{2}{c}{\textbf{Overall}} \\
        \cmidrule(lr){2-3} \cmidrule(lr){4-5} \cmidrule(lr){6-7}
        \cmidrule(lr){8-9} \cmidrule(lr){10-11} \cmidrule(lr){12-13} \cmidrule(lr){14-15}
        & \textit{Acc} & \textit{LEN} &
          \textit{Acc} & \textit{LEN} &
          \textit{Acc} & \textit{LEN} &
          \textit{Acc} & \textit{LEN} &
          \textit{Acc} & \textit{LEN} &
          \textit{Acc} & \textit{LEN} &
          \textit{Acc} & \textit{LEN} \\
        \midrule
        SFT & 13.54 & 7254 & \underline{16.78} & 7011 & 50.23 & 5170 &  72.6 &  3354 & 26.8 & 4742 & 35.7 & 5503 & 35.94 & 5505 \\
        GRPO & 9.9  & 2327 & 8.9  & 2344  & 45.7 & 1540  & 71.5   &  1300 & 28.7  &  1205 & 27.1  & 1840  &   31.96&  1759 \\
        SFT+GRPO & \underline{16.6}  & 4474  & \textbf{18.2}  & 3878  & \textbf{58.4}  & 2783  &  \underline{78.6} &  2124 & 33.8  &  3684 & 37.9  &  3866 &  \underline{40.58} &  3468 \\
        MixPolicy & 14.05  &  2357 & 10.73  & 2001  & 53.04  & 1606  & 75.4  & 1074  & 33.5  & 1129  &  35.7 &  1711 &  37.06 & 1646  \\ 
         \midrule
        \multicolumn{12}{l}{\small{Unified paradigm training}} \\ 
        LUFFY &  15.4 & 3717  & 13.3 & 3017  & 57.34 & 2408  & 78.3 &   1982 &  33.5 & 2216  & \underline{40.6}  & 2614  & 39.74  &  2659 \\
        SRFT{*} & 13.7  & 3344 &  12.1 & 2575  & 53.4  &  2132 &  75.8 & 1650  & 34.3  &  2113 &  36.9 & 2557  & 37.7  & 2395  \\ 
        \midrule
        \multicolumn{12}{l}{\small{Stage-wise training}} \\ 
        ReLIFT & 13.1 & 2236 & 8.9 & 1912 & 50.0 & 1416 & 73.6 & 1086 & 24.6 & 1702 & 34.7 & 1605 & 34.15 & 1659 \\
        BREAD{*} & 13.3  & 2675 & 10.8  & 2034  &  53.6 & 1795  &  74.4 &  1338 & 30.5  &  1752 &  36.1 & 2467  &  36.45 &  2010 \\ 
        \midrule
        Red with (I)& 14.97  & 2288  & 12.45  &  1976 &  52.88 & 1612  & 76.2  &  1053 & \underline{35.2} & 1070 &  36.7  & 1727  & 38.06  &  1621 \\ 
        Red with (II)& 15.73  &  2830 &  13.81 &  2762 & 55.32  &  2007 & 78.3  & 1808  &  30.8 &  1338 & 37.9  & 2309  & 38.64  &   2175 \\ 
        Red(ALL) & \textbf{16.8}  &  3664 & 16.2  & 2896 & \underline{57.5}  & 2047 &  \textbf{80.8}  & 1773  & \textbf{37.6}  & 2226 &   \textbf{40.7} &  2433 &  \textbf{41.6} &  2506 \\ 
        \bottomrule
    \end{tabular}
    \caption{Overall performance on five competition-level mathematical reasoning benchmarks based on Qwen2.5-Math-1.5B. Bold and underline represent the 1st and 2nd in performance. Models marked with an asterisk (*) do not have publicly available code; their results were reproduced based on their respective papers.}
    \label{tab:model_performance}
    
\end{table*}

\section{Experiments}

\noindent{\textbf{Dataset.  }}For the training dataset, we used OpenR1-Math-46k-8192 as provided by LUFFY~\cite{yan2025learning}, with prompts sourced from NuminaMath 1.5~\cite{li2024numinamath} and detailed demonstrations generated by DeepseekR1~\cite{guo2025deepseek}. All trajectory lengths were below 8192 tokens. For the test dataset, we selected MATH500, AIME24/25, AMC~\cite{li2024numinamath}, Olympiad~\cite{he2024olympiadbench}, and Minerva~\cite{lewkowycz2022solving}.

\noindent{\textbf{Evaluation.  }} For our own models, due to the relatively small test sets for AIME 24/25 and AMC, we report avg@32 for these datasets, with a temperature of 0.6. For MATH500, Minerva, and Olympiad, we used pass@1 as the evaluation metric.

\noindent{\textbf{Baselines.  }} We chose Qwen2.5-1.5B-MATH as our base small model for training. The models we compared against include:(1) SFT: Only performs SFT on the dataset.(2) GRPO: Only uses simple GRPO on the dataset.(3) SFT+GRPO: First distills with SFT, then trains with GRPO.(4) GRPO-mixpolicy: Simply merges offline data into the policy optimization function.(5) LUFFY~\cite{yan2025learning}: A shaping mixed-policy GRPO approach.(6) ReLIFT~\cite{ma2025learning}: Interleaves RFT and SFT, with the SFT focusing on the problems that RFT finds hard to solve.(7) SRFT~\cite{fu2025srft}: Conducted based on the entropy of the model's RL stage.(8) BREAD~\cite{zhang2025bread}: Introduces branch rollout to coordinate the two-stage training.

\noindent{\textbf{Implementation Details.  }} For our models, following the LUFFY setup, we adopted a learning rate of 5e-5 for SFT training and trained for 3 epochs. For RL training, we used a learning rate of 2e-6, a temperature of 1.0, and performed 8 rollout iterations. Additional training details are provided in the Appendix.

\subsection{Experimental Results}

\noindent{\textbf{Main Results. }}
Our overall experimental results are presented in Table 2. Our key observations and conclusions are as follows:
(1) Under the more constrained small model setting, SFT models achieve stronger performance compared to purely RL models, but there is considerable room for improvement in token output efficiency.
(2) In most cases, the combination of SFT+RL demonstrates a complementary relationship, effectively balancing model effectiveness and efficiency.
(3) Our experimental comparisons reveal that unified training paradigms, such as LUFFY, SRFT, and RED, exhibit performance advantages over methods that separate SFT/RL training into distinct stages, like ReLIFT and BREAD. This observation further validates the fundamental reliability of our experimental design's starting point.
(4) Our method, \textbf{Red}, demonstrates significantly superior performance when compared to various other methods, proving its effectiveness.

\noindent{\textbf{Ablation Results. }}We conducted an ablation study to assess the effectiveness of each component within our Red framework. As detailed in Table~\ref{tab:model_performance}, we evaluated the impact of two key entropy-aware weighting mechanisms:
(I) Balancing Recall and Extend via Dynamic Entropy Regulation. (II) Accuracy-aware Offline Dynamic Policy Shifts. 

Our results clearly indicate that each component plays a crucial role. Notably, the Dynamic Entropy Regulation(I) achieves significant effects only when combined with the Dynamic Policy Shifts(II). This aligns with our design logic, as effectively balancing SFT and RL requires the initial integration provided by a more unified SFT/RL training context.

\noindent{\textbf{Reasoning Efficiency. }}
In addition to accuracy, we assess the efficiency of different models by measuring the average response token length (LEN) across all benchmarks. Surprisingly, despite their inferior performance relative to LUFFY and SRFT—which employ a unified training paradigm—the two-stage models ReLIFT and BREAD outperform LUFFY in the length of generated solution responses. We hypothesize that this phenomenon arises because the targeted design of the two-stage training approach reduces its susceptibility to offline-sft. While these two solution paradigms each possess distinct advantages, our RED  integrates the strengths of both, exhibiting superiority in both Reasoning Efficiency and performance.

\begin{figure*}[t]
    \centering
    \includegraphics[width=1\linewidth]{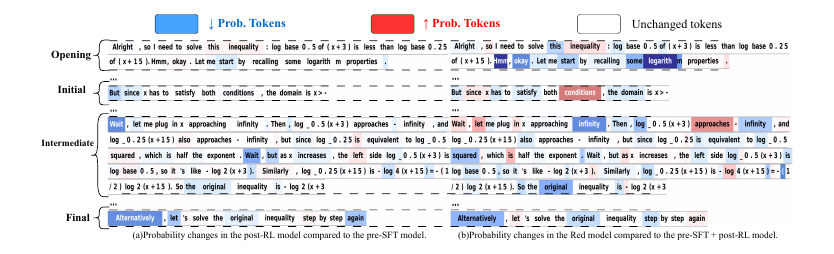}
    \caption{Blue indicates a decrease in probability, red indicates an increase, and white indicates no change, with darker colors signifying a greater degree of change. The reasoning process can be roughly divided into four stages. \textbf{Open}: Typically consists of the model's verbalized preamble or introductory statements. \textbf{Initial}: Located at the beginning of the reasoning process. \textbf{Intermediate}: Constitutes the main body or middle part of the reasoning. \textbf{Final}: Serves as the conclusion or ending of the reasoning.}
    \label{fig:figure5}
\end{figure*}

\subsection{Case study}

To further investigate the differences between the RED model and baseline RL/SFT models, we visualized the probability distribution of the models's token generation during the reasoning process. For comparison, we subtracted the probabilities of two related models to observe the changes induced by training. Blue indicates a decrease in probability, red indicates an increase, and white indicates no change, with darker colors signifying a greater degree of change.

As shown in Figure 5, we illustrate the changes probability  in the RED model compared to the pre-SFT+post-RL model, and the changes in the post-RL model compared to the pre-SFT model. In Figure 5(a), we observe that after RL training, the pre-SFT model shows a decrease in the probability of thinking-related terms (such as ``But", ``Wait", ``Alternatively") across all four stages of reasoning: open, initial, intermediate, and final. While this reduction  might to some extent shorten the length of the reasoning process, it appears to have a negative impact on initiating thoughts (i.e., the ``initial",``intermediate" stage).

In contrast, in Figure5(b) our RED model demonstrates an increase in thinking-related probabilities during the initial and intermediate stages, while showing a decrease in the final stage. This pattern of change aligns more closely with an efficient reasoning process: thorough exploration and consideration during the early and middle stages of reasoning lead to a more comprehensive understanding and judgment, ultimately reducing unnecessary redundant thinking in the concluding phase and directly arriving at a conclusion.

This further highlights the effectiveness of the Red model. The RED model enables deeper contemplation during critical stages and more decisive conclusions, thereby improving both the efficiency and quality of reasoning.

More detailed information regarding the probability changes across different training methods and models in the Appendix.

\begin{figure}
\centering
\includegraphics[width=1\linewidth]{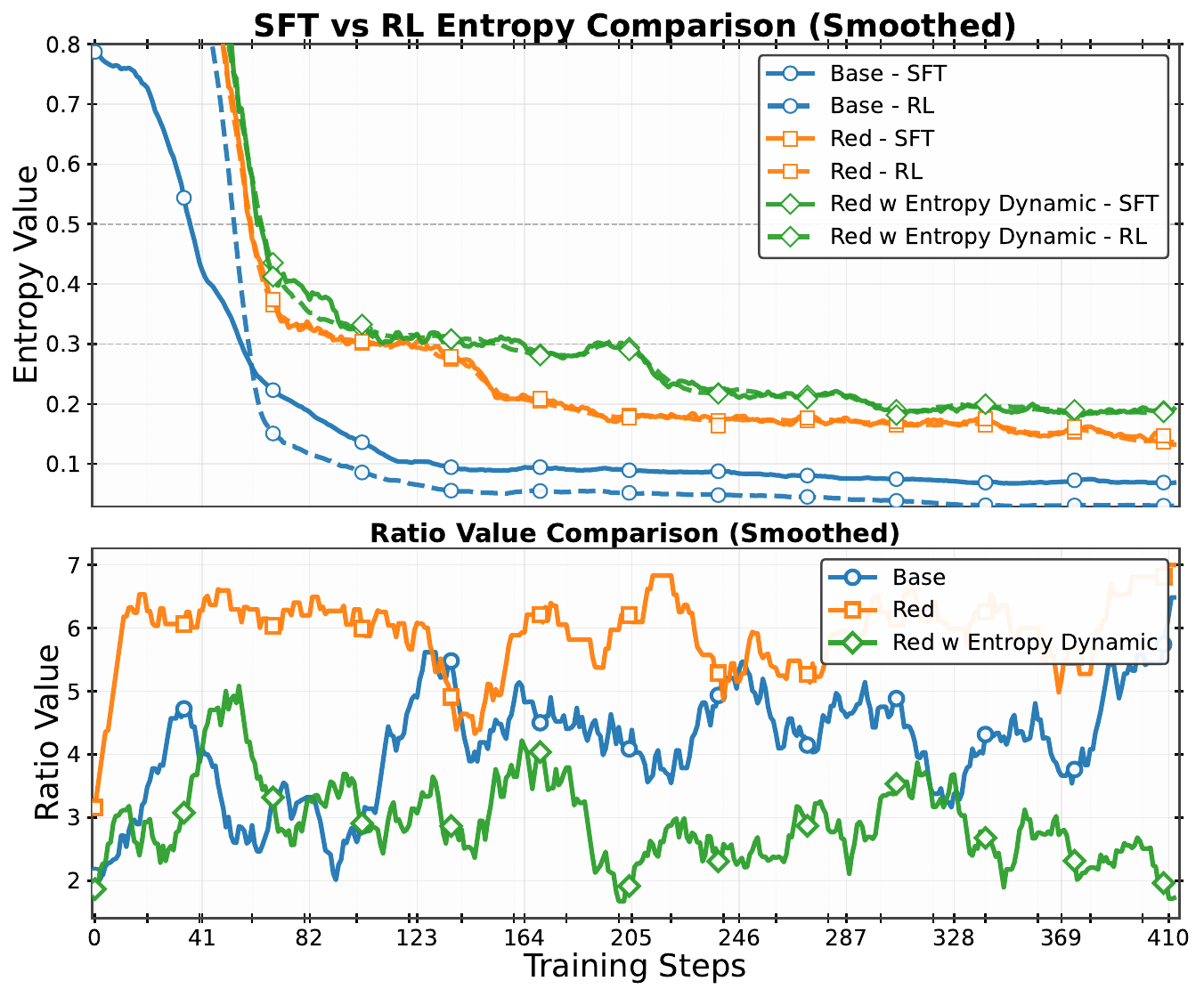}
\caption{Training dynamics during different training setting , including training sft/rl entropy, ratio (w = $\Delta H^{rl}$/$ \Delta H^{sft}$).}
\label{fig:figure6}
\end{figure}

\subsection{Training Dynamics.}

To better illustrate how RED's training progresses, we've visualized the changes in both SFT and RL entropies as well as their changes in ratio (w = $\Delta H^{rl}$/$ \Delta H^{sft}$) in Figure 6.

Figure 6(a) clearly shows that when we apply Accuracy-aware Policy Shifts, both the RL and SFT entropies not only increase significantly and remain at a relatively high level, but also are nearly identical in magnitude. This is a notable improvement compared to simpler approaches like setting $\pi_{\text{offline}}=1$, $\pi_{\text{offline}}=\pi$. Higher entropy generally indicates a greater diversity in the model's policy, allowing for more exploration and less deterministic behavior. 

Furthermore, Figure 6(b) highlights the impact of Dynamic Entropy Regulation. After applying Accuracy-aware Policy Shifts, although the entropy of SFT and RL is effectively adjusted, the fluctuation of their ratio increases and remains relatively high. Through the implementation of Dynamic Entropy Regulation, the balance between RL and SFT is dynamically and more stably maintained. This is crucial as it ensures RL retains the necessary exploration space to discover novel strategies while SFT continues to expand the model's external knowledge, thereby enhancing the robustness and adaptability of the RED training process.

\section{Conclusion}

We introduce \textit{\underline{R}}ecall-\textit{\underline{E}}xtend \textit{\underline{D}}ynamics, a novel method designed to significantly enhance the reasoning capabilities of small language models. It cleverly blends SFT and RL by adaptively balancing when to ``extend" knowledge and when to ``recall" and refine it based on model ``exploration". RED also intelligently integrates distilled data from larger models, allowing SLMs to learn robustly and efficiently. Ultimately, empirical results consistently show that RED outperforms various state-of-the-art methods across multiple mathematical reasoning benchmarks. This not only validates RED's efficacy in boosting SLMs' performance and reasoning efficiency but also marks a significant step forward in developing more capable and practical small-scale models.

\bibliography{aaai2026}

\end{document}